%% file: main.tex
\documentclass[letterpaper, 10 pt, conference]{ieeeconf}  

\usepackage{hyperref}
\usepackage{amsmath} 
\usepackage{amssymb}
\usepackage{graphicx} 
\usepackage{acronyms}
\usepackage{misc}
\usepackage{mfirstuc}
\usepackage{amsfonts}
\usepackage{xcolor} 
\usepackage{comment}
\usepackage{afterpage}
\usepackage{multirow}
\usepackage{comment}
\usepackage{algorithm}
\usepackage{algpseudocode}
\usepackage{float}
\usepackage[figurename=Fig.]{caption}
\usepackage{booktabs}
\usepackage{tabularray}
\usepackage{svg}
\usepackage{subcaption}
\usepackage[symbol]{footmisc} 
\usepackage[noadjust]{cite}

\IEEEoverridecommandlockouts                              

\title{\LARGE \bf
\capitalisewords{Solving Short-Term Relocalization Problems in Monocular Keyframe Visual SLAM using Spatial and Semantic Data}}

\author{Azmyin Md. Kamal$^{1}$, Nenyi Kweku Nkensen Dadson$^{1}$, Donovan Gegg$^{1}$, Corina Barbalata$^{1, *}$
\thanks{*Corresponding author}
\thanks{$^{1}$The authors are with the Department of Mechanical \& Industrial Engineering,  at Louisiana State University, Baton Rouge, LA 70803, USA
        {\tt\small akamal4@lsu.edu, ndadso1@lsu.edu, dgegg1@lsu.edu, cbarbalata@lsu.edu}}%
}

\input{tables.tex}

\begin{document}

\maketitle
\thispagestyle{empty}
\pagestyle{empty}


\begin{abstract}
In \ac{MKVSLAM} frameworks, when incremental position tracking fails, global pose has to be recovered in a short-time window, also known as short-term relocalization.
This capability is crucial for mobile robots to have reliable navigation, build accurate maps, and have precise behaviors around human collaborators.
This paper focuses on the development of robust short-term relocalization capabilities for mobile robots using a monocular camera system. A novel multimodal keyframe descriptor is introduced, that 
contains semantic information of objects detected in the environment and the spatial information of the camera. Using this descriptor, a new \acf{KPR} method is proposed that is formulated as a multi-stage keyframe filtering algorithm, leading to a new relocalization pipeline for \ac{MKVSLAM} systems.
The proposed approach is evaluated over several indoor GPS denied datasets and demonstrates accurate pose recovery, in comparison to a bag-of-words approach.
\end{abstract}

\input{sections/Introduction}
\input{sections/Related_Work}
\input{sections/Methodology}
\input{sections/Experimental_Results}
\input{sections/Conclusions}
\section*{ACKNOWLEDGMENT}
The authors  would  like to acknowledge  financial support  from  NSF $\#2024795$, 
and  the Louisiana Board Of Regents Support Fund, under the Louisiana Materials Design Alliance (LAMDA), provided by the Board as cost share to the NSF under grant number OIA-$\#1946231$.


\vspace{-5pt}
\bibliography{library} 
\bibliographystyle{IEEEtran}

\end{document}

%% file: tables.tex
\newcommand{\neuralnettableNEW}{
    \begin{table*}[t]
    \centering
    \caption{Description of the experimental dataset showing associated platforms, sequences used for testing relocalization methods, sequences used to train their respective YOLOv5L networks and the number of classes detectable.}
    \label{table:dataset}
    \resizebox{\linewidth}{!}{%
    \begin{tblr}{
    vlines, rowsep=0.5pt,
    cell{1-5}{1-5}={halign=c},
    }
    \hline
        Dataset (abbrev.) & Platform & {Sequences used for testing \\ Relocalization methods} & {Sequences used to train \\ Object detector} & {Detectable \\ classes}\\
        \hline
        Machine Hall (MH)&\SetCell[r=2]{c}Drone&MH01, MH02, MH03, MH04, MH05 & MH01, MH04&26 \\
        Vicon Room (VicR)&&V101,V102, V103, V201, V202, V203&V102, V203&15 \\
    \hline
        FR2 Pioneer SLAM (FR2PS)&\SetCell[r=2]{c}Rover&FR2360, FR2PS1, FR2PS2, FR2PS3&FR2PS1, FR2PS2&17\\
        LSU-iCORE-Mono (LiM)&&ROBO0, ROBO1, ROBO2&ROBO0, ROBO1, ROBO2&17\\     
    \hline
    \end{tblr}
    }
    \end{table*}
}

\newcommand{\reloctableNEWFinal}{
    \begin{table*}[t]
    \centering
    \caption{\small Performance of the proposed \ac{PCB} relocalization in comparison to \ac{DBoW2} relocalization. Only for sequences that had ground-truth data, the \ac{ATE} score and the number of frames matched to ground-truth are reported. Last row shows the median value calculated from the averages obtained in the $10$ experimental runs. 
    }
    \label{table:res-relocv2}
    \resizebox{\linewidth}{!}{
    \begin{tblr}{vlines, rowsep=0.5pt,
    cell{1-21}{1-11}={halign=c},
    }
    
    \hline
        Method&\SetCell[c=5]{c}DboW2 Relocalization Module& & & & & \SetCell[c=5]{c}PCB Relocalization Module\\
        \hline
        Sequence&{Avg. \\time (ms)}&{Avg. timesteps \\in lost state (int)}&{Avg. local \\maps (int)}&{ATE \\(m)}&{Frames retained \\for pose estimation \\(int)}&{Avg. \\time (ms)}&{Avg. timesteps \\in lost state (int)}&{Avg. local \\maps (int)}&{ATE \\(m)}&{Frames retained \\for pose estimation \\(int)}\\
    \hline
        MH01&3.208&12&3&0.357&3483&\textbf{2.774}&\textbf{2}&\textbf{1}&\textbf{0.051}&\textbf{3642}\\
        MH02&2.936&10&3&0.201&2843&\textbf{2.592}&\textbf{2}&\textbf{1}&\textbf{0.042}&\textbf{2901}\\
        MH03&\textbf{3.570}&13&4&0.952&\textbf{2610}&4.692&\textbf{4}&\textbf{2}&\textbf{0.950}&2609\\
        MH04&\textbf{2.957}&11&3&0.553&1854&4.551&\textbf{3}&\textbf{2}&\textbf{0.191}&\textbf{1948}\\
        MH05&\textbf{3.907}&12&4&0.526&2070&4.239&\textbf{3}&\textbf{1}&\textbf{0.451}&\textbf{2169}\\
    \hline
        V101&\textbf{2.868}&7&2&0.146&2784&4.486&\textbf{4}&2&\textbf{0.166}&2784\\
        V102&\textbf{3.265}&10&3&\textbf{0.111}&1505&3.455&\textbf{7}&\textbf{2}&0.131&\textbf{1587}\\
        V103&\textbf{3.233}&13&4&\textbf{0.242}&\textbf{1877}&5.402&\textbf{8}&\textbf{3}&0.251&1872\\
        V201&\textbf{2.779}&7&2&\textbf{0.068}&2080&4.737&7&2&0.077&\textbf{2084}\\
        V202&\textbf{3.545}&10&3&0.078&2111&3.386&\textbf{7}&\textbf{2}&\textbf{0.068}&\textbf{2156}\\
        V203&\textbf{3.093}&13&6&0.390&\textbf{1650}&4.193&\textbf{11}&6&\textbf{0.229}&1449\\
    \hline
        FR2360&\textbf{1.852}&14&3&0.481&1052&2.631&\textbf{12}&3&\textbf{0.383}&\textbf{1056}\\
        FR2PS1&\textbf{2.819}&16&8&\textbf{0.346}&1152&2.958&\textbf{13}&\textbf{7}&0.347&\textbf{1295}\\
        FR2PS2&\textbf{2.683}&\textbf{13}&5&0.154&\textbf{634}&2.745&15&5&\textbf{0.076}&510\\
        FR2PS3&\textbf{2.804}&17&5&--&--&3.315&\textbf{15}&5&--&--\\
    \hline
        ROBO0&\textbf{2.885}&13&5&N/A&N/A&4.053&\textbf{9}&\textbf{4}&N/A&N/A\\
        ROBO1&\textbf{3.065}&11&4&N/A&N/A&4.085&\textbf{6}&\textbf{2}&N/A&N/A\\
        ROBO2&3.422&14&6&N/A&N/A&3.422&\textbf{9}&\textbf{4}&N/A&N/A\\
    \hline
        MEDIAN&\textbf{3.011}&13&4&0.294&1973&3.817&\textbf{7}&\textbf{2}&\textbf{0.178}&\textbf{2016}\\
    \hline
    \end{tblr}
    }
    \end{table*}
}


%% file: sections/Introduction.tex
\section{Introduction} 
\label{sec:introduction}

\acf{MKVSLAM} is an optimization-based framework used in mobile robotics. This approach computes the robot's pose while simultaneously creating a sparse-reconstruction of the environment incrementally for each keyframe. 
A keyframe contains compact description of the observed space (no actual image is stored) and is interlinked with others in a directed graph called a pose graph. 
When incremental tracking is interrupted, the global pose must be recovered in a short period of time to maintain the accuracy of the robot's localization and the map usability. In this paper, this event is defined as the short-term relocalization problem.

An important step in recovering global position, in this framework, is to choose a number of appropriate keyframes from the pose graph. Additionally, information that describes the physical space where incremental tracking was interrupted, is useful for subsequent pose recovery  \cite{fast-reloc-og}. This step is defined as the \acf{KPR}. In this regard, each keyframe requires a unique description that differentiates it from other keyframes in the pose graph.

Several \ac{MKVSLAM} frameworks choose keyframes for pose recovery using the \ac{DBoW2}  method \cite{lopez2012dbow2}. In this method, keyframes are described using collections of non-semantic local 
features.
However, the bag-of-words method has shown limited success for short-term relocalization since the events that trigger tracking loss decrease the number of local features identified. Furthermore, \ac{DBoW2} does not contain spatial data~\cite{lowry2015vprsurvey}.

Some \ac{MKVSLAM} frameworks have integrated human-understandable semantics into their pipeline. 
In these works, semantic data offers additional information for geometric landmarks such as points, planes, and objects~~\cite{zhou2022point-plane-object-slam} and can provide reliable solutions for people 
tracking~\cite{younes2017}.
%

\begin{figure}[t]
\centering
     \includegraphics[width=\columnwidth]{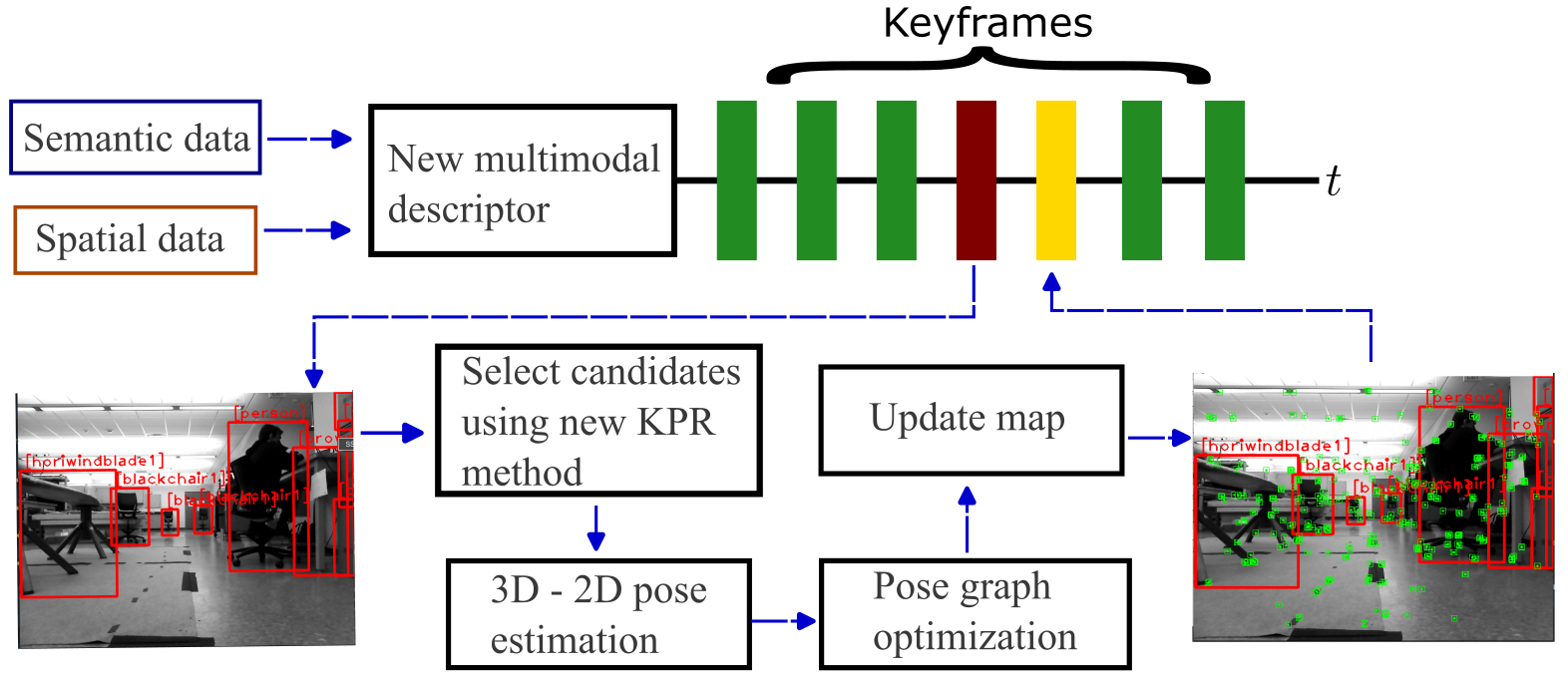} 
      \caption{{\small{High level system overview: For each keyframe (colored rectangles) the proposed multimodal descriptor is formed using semantic and spatial data. When tracking is lost in the red keyframe, the proposed \ac{KPR} method selects a number of keyframes from the pose graph. 
      By solving a series of correspondences between 3D map points (in a candidate keyframe) to 2D keypoints (in query keyframe), an estimation of the global pose is found. Followed by a pose graph optimization step, global pose is recovered in the yellow keyframe and incremental tracking resumes in green keyframes. 
      }}}
       \label{fig:intro}
\end{figure}

This paper leverages the advancements in semantic data extraction using deep neural networks and presents a fast and accurate solution to the short-term relocalization problem. The proposed approach introduces a novel multimodal keyframe descriptor consisting of semantics and spatial data from monocular images. This is integrated into a 
\ac{KPR}
method that chooses appropriate keyframes candidates by passing them through a multi-stage filtering algorithm. 
The novel \ac{KPR} is coupled with an off-the-shelf 3D-2D pose estimation technique to create a novel relocalization pipeline that is 
shown in \figurename~\ref{fig:intro}. 
By incorporating the novel relocalization method into ORB-SLAM3~\cite{orb-slam-3}, an open-source \ac{MKVSLAM} framework, the robustness and efficiency of the proposed approach is demonstrated for aerial and ground robot datasets operating in GPS-denied environments.

The contributions of this paper are as follows:
\begin{itemize}
  \item A new keyframe descriptor called the \acf{PSD} is proposed. It utilizes semantic data and camera pose to uniquely characterize keyframe objects in the pose graph. Using this descriptor, a novel \ac{KPR} algorithm called the \acf{PCB} is formulated.
  \item The integration of the proposed \ac{KPR} method for improving short-term relocalization within the ORB-SLAM3 \ac{MKVSLAM} framework. 
\end{itemize}

The remainder of the paper is organized as follows: Section~\ref{sec:related-work} discusses related keyframe descriptors and global pose recovery methods based on keyframes. Section~\ref{sec:methodology} presents the proposed \ac{PSD} descriptor, the \ac{KPR} method, and how they are integrated within a \ac{MKVSLAM} system. In Section~\ref{sec:results}, the experimental results are discussed. Section~\ref{sec:conclusions} summarizes findings and discusses future work.

%% file: sections/Related_Work.tex
\section{Related Work}
\label{sec:related-work} 

\begin{figure*}[t]
\begin{center}
    \includegraphics[width=0.85\textwidth]{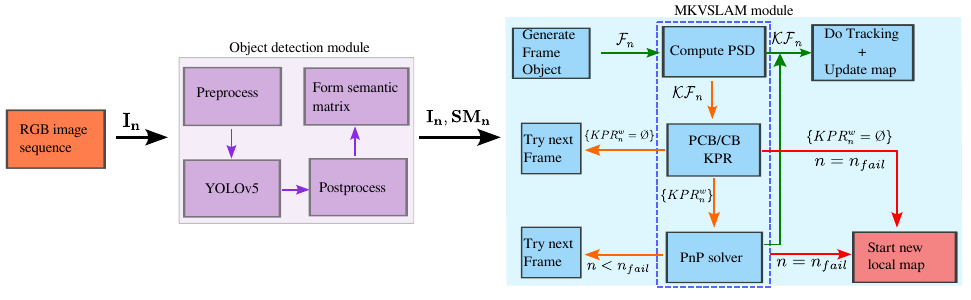}  
\end{center}
    \caption{\small{Overview of the full framework:
    An image is passed through the purple block where the \textit{preprocess} block is resizing the image, that is next fed to the YOLOv5 object detector~\cite{yolov5}. Next the \textit{postprocess} block performs non-maximal suppression~\cite{neubeck2006efficient} 
    on the detection to finalize class predictions for the detected objects. In light blue the \ac{MKVSLAM} module is presented. The path in green shows the active tracking state. The path in orange shows a relocalization attempt in a lost state. The path in red indicates a full recovery failure. The relocalization pipeline identified by the dashed deep blue rectangle consists of computing the \ac{PSD} descriptor of the query keyframe, choosing optimal candidates using the proposed \ac{KPR} method and recovering global camera position by solving a series of \ac{PnP} problems. 
    }}
    \label{fig:sys_overview}
\end{figure*}

\underline{\textit{Descriptors for keyframes:}}
A variety of \ac{MKVSLAM} frameworks using local feature~\cite{orb-slam-3}, feature learning~\cite{yang2019keyframe}, or object detection~\cite{yang2019cubeslam} 
have adopted the visual bag-of-words model to describe keyframes. This descriptor is created by extracting local features using algorithms such as \ac{ORB}~\cite{orb-paper} or \ac{SIFT}~\cite{lowe1999sift}, and aggregating those features into a collection of visual words based on a pre-trained vocabulary~\cite{lopez2012dbow2}. While it is pose-invariant, efficient in traversing the pose graph, and chooses optimal keyframes for loop closures~\cite{fast-reloc-og}, it lacks any spatial or semantic data. 
This additional information has been shown to improve localization when there is a large viewpoint difference, low texture, or noise~\cite{lowry2015vprsurvey}. 
Neural network based approaches aim to learn on how to extract local features~\cite{lift-slam}, update the vocabulary online ~\cite{garcia2018ibow}, and compute descriptors as a function of the image~\cite{kuse2021learning}.

\underline{\textit{Relocalization approaches:}} A fast global pose recovery method was introduced in \cite{fast-reloc-og} where keyframes were uniquely identified using the bag-of-words model. Here non-semantic local features were aggregated into 
visual words~\cite{video-google} 
using a pretrained dictionary. 
When tracking fails, a \ac{KPR} method retrieved a number of candidate keyframes based on a normalized L1 score between the query and candidate keyframes. No semantic data was considered in this approach. 
Bruno and Colombini integrated the \ac{LIFT} \cite{li-lift} with ORB-SLAM1 
to form a hybrid monocular \ac{VSLAM} system \cite{lift-slam}. A drone and an outdoor ground robot dataset were used in their testing. The approach demonstrated higher positional accuracy when compared with ORB-SLAM1. However, this method focused on the feature extraction and still relied on the bag-of-words approach for relocalization. 
In \cite{improve-reloc-vr}, class labels of objects were detected using YOLO bounding-box regressor \cite{yolo} and were used to form an object label array. An \textit{object similarity} score was computed by comparing the class label arrays between query and candidate keyframes to perform place recognition. No spatial information (pose, bounding box) was used, and the size of the object label array is needed to be known apriori.

\underline{\textit{Semantics for localization: }}
%
In \cite{sem-map-invariant} a metric map was augmented with the 3D poses of objects detected using a neural network. Here, relocalization was demonstrated as a long-term data association problem where global pose was recovered by aligning two semantic maps recorded for the same environment but taken at two different instances.  
%
In \cite{vslam-human} two approaches for accurate pose estimation in human populated indoor environments were presented. Using YOLO \cite{yolo} to detect objects and Mask-RCNN \cite{mask-rcnn} to perform semantic segmentation, the approach was integrated with ORB-SLAM2 \ac{MKVSLAM} framework.  
%
Another work building on top of ORB-SLAM2 was proposed in \cite{yang2019keyframe} where semantic segmentation data and a geometric constraint were used to develop a new local keypoints extraction method. This method filtered out keypoints belonging to moving objects to improve position estimation accuracy and the semantic data was used create a dense metric-semantic map. 
%
%
Similar to the above, in DynaSLAM \cite{bescos2018dynaslam} and DyOb-SLAM~\cite{wadud2022dyob}, moving objects were semantically segmented to remove the keypoints associated with them. This was shown to jointly improve camera pose estimation, and tracking position of mobile objects in the active map. 

%% file: sections/Methodology.tex
\section{Methodology}
\label{sec:methodology}


Figure~\ref{fig:sys_overview} shows the framework for the short-term relocalization approach. 
The method uses an Object Detection module to extract semantic data, shown in light purple, based on a \ac{DNN} architecture. 
This data is passed into a \ac{MKVSLAM} system (shown in light blue). In the following paragraphs, the new keyframe descriptor is presented and the keyframe-based place recognition approach is detailed. 
\subsection{Pose Semantic Descriptor in the context of \ac{MKVSLAM}} \label{psd}
The proposed approach starts by extracting meaningful information from the environment and generates semantic data based on identified objects. That is, at each timestep, $n:n\in \mathbb{Z^+}$, an RGB image $\mathbf{I_n}$ is passed through the Object Detection module that contains a
\ac{DNN} object detector as described in \cite{yolo}. The output is a matrix named here the \textit{semantic matrix} $\mathbf{SM_n} \in \mathbb{R}^{b \times 5}$ where $b \in \mathbb{Z^+} $ is the number of objects detected. 
Each row of this matrix contains the class label and the 2D bounding box coordinate of an object described by the top-left and bottom-right corners in the image coordinate system.
The class labels are defined using an unique integer number. 

A copy of the original image, $\mathbf{I_n}$, and the semantic matrix, $\mathbf{SM_n}$, are passed to the \ac{VSLAM} system (in light blue) were a frame object, \( \mathcal{F}_n = \{{\mathbf{I_n}}, \mathbf{SM_n}, \{ x_n^{z_1} \}, \mathbf{T_{wn}} \}\) is created. Here, $ \{ x_n^{z_1} \} $ are the 2D non-semantic features extracted from the image (keypoints), and $\mathbf{T_{wn}}$ is an initial estimate of the camera pose represented in $SE(3)$.
$\mathbf{{T}_{wn}}$ is refined over time using \ac{BA} 
\cite{younes2017}. 

The SLAM module has modes of operations (referred to as states) as depicted with colored arrows in Figure~\ref{fig:sys_overview}. The green path, defined as a tracking state, refers to the continuous simultaneous estimation of the camera pose and environment mapping. The orange path, defined as a lost state, refers to the state where the system attempts to recover from global position loss. This recovery must happen within a certain number of time steps ($n_{fail}$) from the moment that tracking loss occurs.    
If the global pose is not recovered within $n_{fail}$ time steps, the systems follows the red path defined as a failure state. In the failure state, the current map closes and a new map is initialized. Each active map denoted by $\mathcal{M^*}$ is defined as a local map. One session may contain multiple local maps but only one is active at any given time. 
In the tracking and lost states, a frame object is used to create a keyframe object 
\( \mathcal{KF}_n = \{{\mathcal{PSD}_n, \mathbf{X_n}}\}\), 
where \( \mathbf{X}_n \in \mathbb{R}^{3} \) are the initial position estimates of 3D map points associated with $ \{ x_n^{z_1} \} $ keypoints. If no objects are detected, then $\mathcal{PSD}_n,$ is not generated and no keyframe is created from frame $\mathcal{F}_n,$.

To increase the knowledge of the system in defining a keyframe object for fast and accurate candidate selection, this paper presents the novel multimodal keyframe descriptor called the \acf{PSD}. The proposed descriptor is formulated as the following data structure:
\begin{equation}\label{eqn:psd_data}
  \begin{gathered}
    \mathcal{PSD}_n = \left \{ n, \left \{ sl_n^k \right \}, \mathbf{B_n}, \mathbf{{T}_{wn}} \right \}\
  \end{gathered}
\end{equation}%
where, $n$ is the timestep, $\left \{ sl_n^k \right \}$ denotes the set of class labels for the $k$ objects detected. $\mathbf{B_n}~\in~\mathbb{R}^{k\times4}$ are their 2D bounding box coordinates, and $\mathbf{{T}_{wn}}$ is the camera pose. The multimodal nature of the descriptor allows the semantic (objects detected and associated class labels) and geometric data (camera pose) to complement each other for place recognition. 

\begin{figure*}[t]
\begin{center}
  \includegraphics[width=0.85\textwidth]{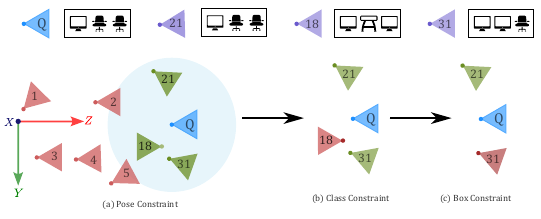}  
\end{center}
  \caption{\small{Application of proposed \ac{KPR} method in a 2D case. The top row shows the collection of objects seen by the query keyframe and some of the candidate keyframes.  
  Colored triangles in active map represents keyframes, with red being identified those that aren't chosen by the proposed approach. Keyframes that are selected in each of the three steps are colored in green. The shaded blue circle is the search sphere in the \textit{pose constraint}. In the \textit{class constraint} only keyframes 21 and 31 are selected as they contain the same type of objects as the query keyframe (identified by blue triangle). The \textit{box constraint} retains only keyframe 21 as the location of the observed objects matches the query keyframe.}}
  \label{fig:pcb_apply}
\end{figure*}
%
\subsection{Pose-Class-Box \acf{KPR} method} \label{sec:pcb_method}
In the following paragraphs a new \ac{KPR} method is presented, named the \textit{Pose-Class-Box} approach. This is activated during a lost state and is formulated as a three-stage filtering algorithm.

In the first stage, spatial data is used to define a \textit{pose constraint}. Consider the pose of the camera represented by $\mathbf{T} \in \mathbb{R}^{4 \times 4}$:
\begin{equation}\label{t_mat}
  \begin{gathered}
    \mathbf{T} = \begin{bmatrix}
                        \mathbf{R} & s.\mathbf{t}\\
                        \mathbf{0} & 1
                 \end{bmatrix}  = \begin{bmatrix}
                                            r_{11} & r_{12} & r_{13} & s.t_{x} \\ 
                                            r_{21} & r_{22} & r_{23} & s.t_{y}\\ 
                                            r_{31} & r_{32} & r_{33} & s.t_{z}\\ 
                                            0 & 0 & 0 & 1 
                                    \end{bmatrix}
  \end{gathered}
\end{equation}
where \( \mathbf{R} \in \mathbb{R}^{3\times3} \in SO(3) \) is the rotation matrix, \( s \in \mathbb{R}\) is the scale factor, and \( {\mathbf{t} = \left [ t_x, t_y, t_z \right ]^T} \in \mathbb{R}^3 \) is camera`s translation in the world coordinate frame $w$. $\mathbf{T}$ is non-singular matrix, independent of the scale factor $s$.

For two camera poses $\left( \mathbf{T_{w1}}, \mathbf{T_{w2}} \right)$ that are similar, the following relation is defined:
\begin{equation} \label{t_new_rel}
\begin{aligned}
      \text{if} \quad  \mathbf{T_{w1}}  \approx & \mathbf{T_{w2}} \Rightarrow   \\ \delta T_{1,2}  = & |\ ||\mathbf{T_{w1}}||_F - ||\mathbf{T_{w2}}||_F\ | \in \mathbb{R^+} \approx 0
\end{aligned}
\end{equation}
where $ || \cdot ||_F$ is the Frobenius norm \cite{fundamental-matrix}. Based on Equation~\eqref{t_new_rel}, in this work the geometrical similarity between two keyframe poses is represented by one real number. This is computed pairwise between the query keyframe and all other keyframes in the pose graph. For a given pair of keyframes at times $n$ and $l$, let this number be denoted $\delta T_{n,l}$ and computed as:
\begin{equation}\label{eqn:frob}
  \begin{gathered}
   \delta T_{n,l} = ||\mathbf{T_{wn}} - \mathbf{T_{wl}}||_F = \left (\sqrt{\sum_{i=1}^{n}\sum_{j=1}^{m} |a_{ij}|^2}  \right )
  \end{gathered}
\end{equation}
Using the above equation, the procedure of selecting keyframe candidates from the pose graph in the active map $\mathcal{M}$ has to satisfy Equation~\eqref{eqn:pose_const}. This is defined as the \textit{pose constraint} step.

\begin{equation}\label{eqn:pose_const}
  \begin{gathered}
   \epsilon \leq \delta T_{n,l} \leq \delta T_{th},\ 
  \end{gathered}
\end{equation}
where $\delta T_{th}$ is a predefined threshold and  $\epsilon=10^{-4}$ is to account for numeric errors. 
The list of candidates after the first filtering stage is marked by $ \left \{ P_n^j \right \}$. 
The \textit{pose constraint} step can be described as selecting candidates within the search sphere centered around the query keyframe $\mathcal{KF}_n$ with a radius $\delta T_{th}$ (shaded-blue circle in \figurename~\ref{fig:pcb_apply}). This keyframe is created by computing its \ac{PSD} and retaining 3D map points $\mathcal{X}_n$ from frame $\mathcal{F}_n$.

Note that, only using the \textit{pose constraint} described above, a candidate is selected from search sphere even if it's orientation is significantly different from $\mathcal{KF}_n$ (blue triangle in \figurename~\ref{fig:pcb_apply}). This can lead to choosing keyframes ill-conditioned to recover camera pose using 3D-2D pose estimation techniques. 
To address this limitation, semantic data is incorporated into the proposed \ac{KPR} approach in the second stage, \textit{class constraint}. This is detailed below. 

Let the list of classes of objects be $\left \{ c_n \right \}$ in the query keyframe and $\left \{ c_j \right \}$ in the $j^{th}$ candidate keyframe. As the class labels are represented by unique integer numbers, a numeric score, $\delta C_{n,j} \in \mathbb{R}^+$, describing the similarity of the class labels between $\left \{ c_n \right \}$ and $\left \{ c_j \right \}$  is computed as:
\begin{equation}\label{dels_nl}
  \begin{gathered}
   \delta C_{n,j} = \left |||\left\{ c_n \right \}||_2 - ||\left\{ c_j \right \}||_2  \right |
  \end{gathered}
\end{equation}
The above numeric score is computed pairwise between the query keyframe and all the $l$ candidates in $ \left \{ P_n^j \right \}$. Each pairwise score is stored in the list $ \left \{ C_n^j \right \}$. A keyframe candidate is considered semantically relevant if the below condition is fulfilled:



%
\begin{equation}\label{eqn:class_const}
  \begin{gathered}
   \delta C_{th} \leq \delta C_{n,j} \leq \delta C_{th} + 0.1 \delta C_{th}
  \end{gathered}
\end{equation}
where $\delta C_{th}$ is an adaptive threshold computed as: 
\begin{equation}\label{eqn:dels_Cth}
  \begin{gathered}
   \delta C_{th} =  \min\left ( \left \{ C_n^j \right \} \right)
  \end{gathered}
\end{equation}
%

The list of candidates after the
second filtering stage is marked by $ \left \{ R_n^j \right \}$. 
This formulation has two advantages. Firstly, it is order independent, i.e the same score is obtained regardless the order the class labels appear in the semantic matrix. Secondly, the number of objects recorded in the two keyframes is not required to be same. 

However, representing classes as integers may lead to some ambiguous cases resulting in wrong keyframe selections. For example, if $\left \{ c_n \right \} = \{1,1,1,1\}$ and $\left \{ c_j \right \} = \{4\}$ then both $\delta C_{th}$ and $\delta C_{n,j}$ are equal leading to a wrong selection.
To address this issue and further reduce the pool of candidates, the last stage of the \ac{KPR} approach is focused on filtering keyframes based on the location of the bounding boxes of the objects in the image coordinate frame, referred as the \textit{box constraint} step.  

If the $j^{th}$ candidate satisfies Equation~\eqref{eqn:dels_Cth}, let $\mathbf{B_n} \in \mathbb{R}^{b_n\times4}$ be the bounding box coordinates of objects recorded in the query keyframe and $\mathbf{B_j} \in \mathbb{R}^{b_j\times4}$ represent the bounding box coordinates of objects recorded in the $j^{th}$ candidate.
The \ac{IoU} scores \cite{giou} between the bounding boxes in the query keyframe and all the candidates in $\{R_n^j\}$ is computed. To establish the objects that are visible both in the query and candidate keyframes, a threshold is used:
\begin{equation}
    \label{eq:th_box}
    \text{if} \quad \%IoU > \delta IoU \Rightarrow v_j = v_j+1
\end{equation}
where $\%IoU$ is the percentage overlap, $\delta IoU$ is a predefined threshold, and $v_j$ is a counter for objects that overlap in the scene. 
If a certain number of objects have been identified as being in both keyframes, the candidate is considered relevant for relocalization. This can be formulated as: 
\begin{equation}\label{eqn:box_const}
  \begin{gathered}
  \text{if} \quad v_j \geq b_n \Rightarrow j^{th} \ \text{keyframe} \in  \left \{ KPR_n^w \right \}
  \end{gathered}
\end{equation}
where $b_n$ is the number of objects in the query keyframe and $ \left \{ KPR_n^w \right \}$ is the final candidate list. 
%


However, if global pose is not recovered at the exact $n^{th}$ timestep, then frames {$\mathcal{F}_{n+1}, \mathcal{F}_{n+2}$, ...} have their camera pose initialized to an identity matrix, $\mathbf{I}_{4 \times 4}$. This is due the fact that in a monocular 
\ac{MKVSLAM} system, no other information beyond image data is available to estimate camera pose. 
In these frames, computing $\delta T_{n,l}$ using Equation~\eqref{eqn:frob} will result in computing the Euclidean distance of $\mathcal{KF}_n$ from origin only, resulting in the \textit{search sphere} pulling candidates that may have no semantic or geometric relevance.
In this case, the \textit{pose constraint} step should be skipped, and only the \textit{class constraint} and \textit{box constraint} steps are used to select candidates from the pose graph in the active map $\mathcal{M}$. This is marked as the \textit{Class-Box} \ac{KPR} method. To select either the PCB or the CB \ac{KPR} methods, the following conditional check is used:
\begin{equation} \label{eqn:nTest}
  \begin{gathered}
   \text{if} \quad \delta T_{L_2} = ||\mathbf{T_{wn}} - \mathbf{I}_4||_2 > 0 \Rightarrow \text{full \ac{PCB}}
  \end{gathered}
\end{equation}

%
\neuralnettableNEW 
%
\subsection{Proposed short-term relocalization method} \label{sec:prop_reloc_method}
The proposed short-term relocalization method is implemented as shown in Algorithm~\ref{pcb-reloc}.
The hyperparameters of the algorithm are $\Delta T_{th}$, $n_{fail}$. 
Line 2 extracts all the keyframes $\{ K_n^i \}$ in the pose graph at $n^{th}$ timestep where $i$ represents the number of keyframes.
Line 5 computes the \ac{PSD} descriptor as expressed in Equation~\eqref{eqn:psd_data}.
Line 9 calls for the full \ac{PCB} method and line 11 calls the reduced version as described by Equation~\eqref{eqn:nTest}.
Line 14 recovers the global position in the active map $\mathcal{M}$ by solving a series of 3D to 2D motion estimation problems using the \ac{MLPnP}~\cite{mlpnp} solver. 
Here $bFlag$ is set to true if a pre-defined number of map points to keypoints correspondences are found.
If $bFlag$ is true, tracking state is resumed (system follows the green path in Figure~\ref{fig:sys_overview}), 
else  Algorithm~\ref{pcb-reloc} is repeated as long as recovery is not successful within $n_{fail}$ timesteps. If this is not achieved, a new local map is initialized, as shown in lines 19-21. 

\begin{algorithm}
\small 
\caption{Relocalization using \ac{PCB}}\label{alg:pcb_alg}
\begin{algorithmic}[1]
\Require Frame $\mathcal{F}_n$, Map $\mathcal{M}$, timestep $n$, $\delta T_{th}, n_{fail} $  
\State $n_r \gets 1$
\State $\{ K_n^i \} \gets \mathcal{M}.GetAllKeyFrames()$
\For{$n_r < n_{fail}$}
        \State $\mathcal{KF}_n \gets KeyFrame\left ( \mathcal{F}_n \right)$ 
        \State $\mathcal{KF}_n.ComputePSD\left (  \right )$
        \For{$j \gets 1$ to $i$}
            \State $\delta T_{L_2} \gets || \mathbf{T_{wj}} - \mathbf{I_4} ||$
            \If{$\delta T_{L_2} > 0$}
                \State $\left \{ KPR_n^w \right \} \gets PCB\left ( \mathcal{KF}_n,\{ K_n^i \}, \delta T_{th} \right )$ 
            \Else{}
                \State $\left \{ KPR_n^w \right \} \gets CB\left ( \mathcal{KF}_n,\{ K_n^i \}, \delta T_{th} \right )$ 
            \EndIf
        
        \EndFor
    \State $bFlag \gets PerformRelocalization\left ( \left \{ KPR_n^w \right \} \right)$ 
    \If{$bFlag == True$}
        \State \textbf{Return} $bFlag, \left \{ KPR_n^w \right \}, \mathcal{KF}_n$
    \Else{}
        \State $n_r \gets n_r + 1$
        \If{$n_r >= n_{fail}$}
            \State \textbf{Stop} \ac{PCB}
            \State \textbf{Terminate} $\mathcal{M}$
        \EndIf
    \EndIf
    
\EndFor
\label{ln:blfag}\label{ln:perform_reloc}
\end{algorithmic}
\label{pcb-reloc}
\end{algorithm}

%% file: sections/Experimental_Results.tex
\section{Experiment \& Results}
\label{sec:results}
This section details the algorithm implementation, experimental setup, and the obtained results. 

\subsection{Implementation details} 
\label{sec:exp_setup}
The proposed approach is implemented and tested on a computer with Ubuntu 22.04, Intel i5-9300H $@$ 2.4GHz, NVIDIA RTX 2060 GPU, and 16 GB RAM. 
The \ac{PSD} keyframe descriptor and \ac{PCB} \ac{KPR} method were 
incorporated in ORB-SLAM3. 
Multi-mapping, loop closure and four parallel threads were used and the proposed modifications targets the relocalization function in the Tracking, Local Mapping, and Keyframe Database modules in the ORB-SLAM3 framework \cite{orb-slam-3}.
For the object detection module, the YOLOv5L has been chosen due to its fast computational capabilities \cite{yolov5}.
The complete system was implemented in \ac{ROS} \cite{macenski2022ros2}. 
The source code and the datasets used for this paper are made publicly available at the link below\footnote{\url{https://github.com/RKinDLab/ros2_psd_pcb_reloc}}.
The parameters for the proposed relocalization method were set as follows: {$n_{fail}$ was set to $20$ timesteps that approximately corresponds to one second, $\delta T_{th} = 0.5$} and  $\delta IoU = 90\%$ were set based on trial and error.

\subsection{Dataset} \label{sec:exp_dataset}
The experiment is performed on \textbf{$18$} indoor image sequences spanning across four unique datasets split into two drone and two ground robots datasets, as shown in Table \ref{table:dataset}. Sequences for the drone dataset are taken from the  
\textit{Euroc MAV} dataset \cite{euroc-dataset}. The sequences present a drone maneuvering in different indoor environments containing variety of objects. This dataset has ground-truth camera poses. 
One ground robot dataset is taken from the \textit{FR2 PIONEER SLAM} sequences in the {\textit{TUM {RGB-D}}} dataset \cite{sturm12iros}. This dataset presents motion-blur in the images and discontinuities in the recording, that can induce the lost state. 
Ground-truth poses are available in this dataset. Another ground vehicle dataset is created for this project and referred as the \textit{LSU iCORE Mono} dataset. This is an indoor dataset representing a robotics laboratory showcasing a cluttered environment, with multiple  chairs, people, workbenches, other robots, whiteboards, etc. This dataset has no ground-truth camera poses. The YOLOv5L network is trained for the Object Detection module for each of the four datasets as shown in Table~\ref{table:dataset}. An NVIDIA RTX 3080 Ti was used to train all the networks.

\subsection{Experiment} \label{sec:exp_experiment}
The proposed relocalization method, as introduced in Section~\ref{sec:prop_reloc_method}, is tested in all the $18$ image sequence described before. In the sequences from \textit{EuRoC MAV} and \textit{LSU-iCORE-Mono} datasets, $5$ lost states are artificially induced at random times. 
As the \textit{FR2 PIONEER SLAM} dataset is prone to lost state occurrences, there was no need to artificially induce the lost state. Each sequence is run $10$ times and a series of quantitative metrics are computed to evaluate the performance of the system. For comparison, the \ac{DBoW2} based relocalization method in ORB-SLAM3 is used as the baseline for analyzing the obtained results. 

\subsection{Metrics} \label{sec:exp_metrics}
The \ac{mAP} metric \cite{pascal-voc} 
is chosen to measure the detection performance of the Object Detection module. 
For evaluating the \ac{KPR} method, the average execution time and the average keyframe candidates selected are recorded. The average total execution time, the average number of frames spent in lost state, and the average number of local maps created when pose recovery fails are used to study the relocalization method. 
To assess the 
impact of extended tracking loss 
\ac{ATE} \cite{sturm12iros} 
score is calculated. 
\ac{ATE} scores are reported only for sequences that have ground-truth camera poses. 


\reloctableNEWFinal

\subsection{Results and Discussion} \label{sec:result_discuss}
Prior to conducting the experiment, it is necessary to assess if the Object Detection Module has high precision. 
This is important to maximize the effectiveness of semantic information within the proposed relocalization method. 
The \ac{mAP} scores are $82.9\%$, $83.5\%$, $84.0\%$ and $91.0\%$ for the \textit{MH}, \textit{VicR}, \textit{FR2PS} and \textit{LiM} datasets respectively. 
The average inference time is $29.55$~ms which is lower than the average time needed by ORB-SLAM3 to process one image \cite{orb-slam-3}.

\begin{figure}[t]
  \includegraphics[width=1.0\columnwidth]{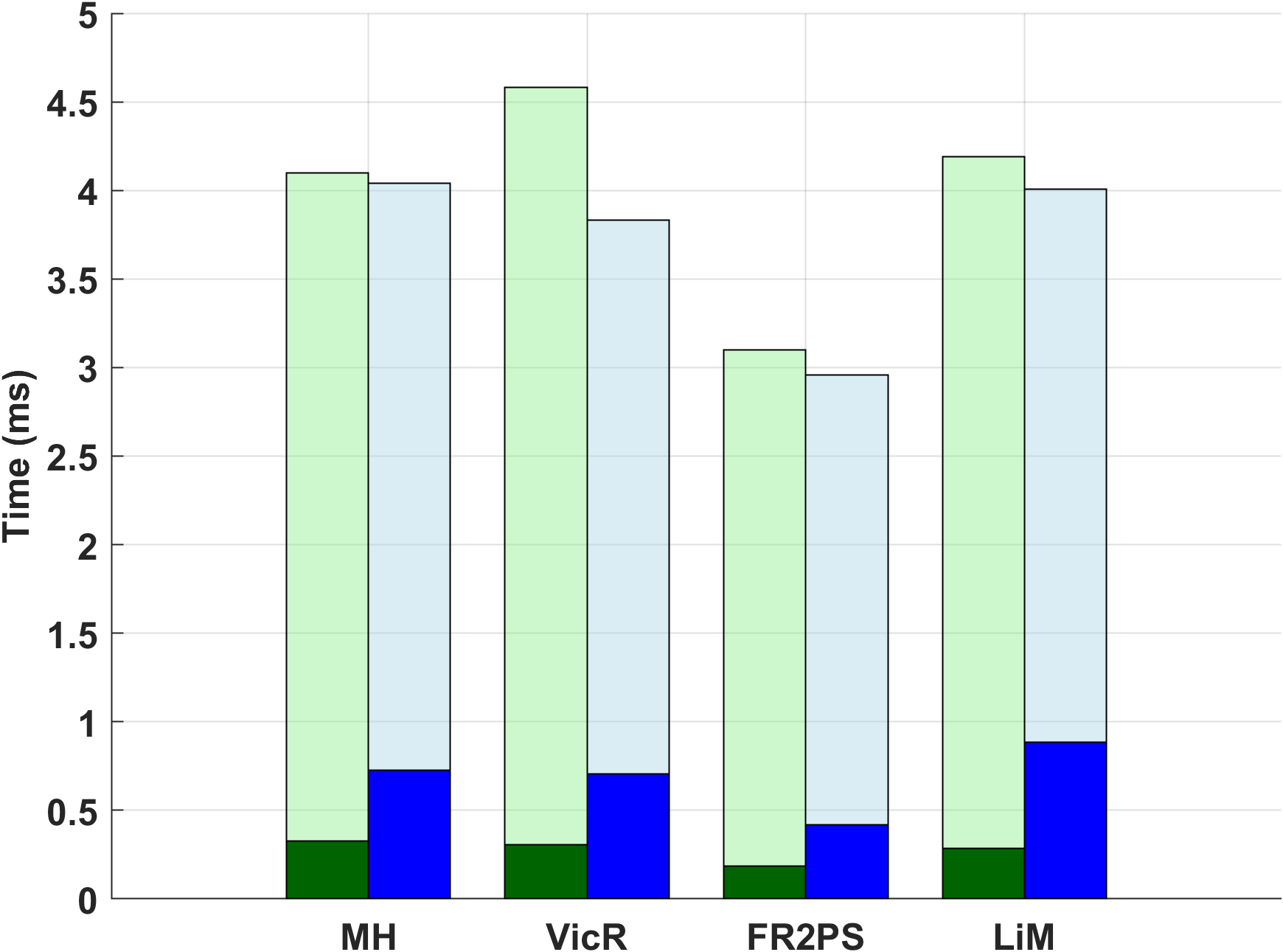}
  \caption{\small{Execution time for the \ac{KPR} methods and the full relocalization method. Deep green bar represents the time for proposed \ac{KPR}, deep blue for \ac{DBoW2} \ac{KPR}, light green for the proposed relocalization method and light blue for the \ac{DBoW2} relocalization method. Best viewed in color.
  }}
  \label{fig:time_bars}
\end{figure}

An important requirement for the proposed multimodal keyframe descriptor is to facilitate fast candidate retrieval when performing relocalization \cite{fast-reloc-og}. As depicted in Figure~\ref{fig:time_bars}, the proposed \ac{KPR} method is faster than the \ac{DBoW2} method in all four datasets with 
an average execution time of $0.297$ ms, being two times faster than the DBoW, that has an average execution time of $0.702$ ms. 
%
The proposed \ac{KPR} approach extracts on average five, six, four, and four keyframe candidates from the \textit{MH}, \textit{VicR}, \textit{FR2PS}, and \textit{LiM} datasets respectively. In comparison, the DBoW2 \ac{KPR} selected, on average, only one candidate in all datasets. 

These results show that the inclusion of both geometrical and semantic information into the keyframe descriptor improves its capabilities by allowing more keyframe candidates to be selected at a faster rate than the \ac{DBoW2} approach.
The performance of the proposed relocalization method is shown in Table~\ref{table:res-relocv2}. The proposed approach has an average execution time of $3.817$ ms, while the \ac{DBoW2} relocalization method has an average execution time of $3.011$ ms. 
Although the proposed \ac{KPR} method demonstrates rapid keyframe selection, in average $0.297$ ms, the main computational load during relocalization falls on the \ac{PnP} solver. This is primarily due to \ac{KPR} generating four times more keyframe candidates than the baseline \ac{DBoW2}, leading to increased processing of 3D map points to 2D key point correspondences. Despite this, the proposed method maintains a real-time speed, approximately $4$ ms.
The proposed relocalization method performs better than the \ac{DBoW2} approach by reducing the time spent in the lost state by approximately $50\%$. 
This is shown in the the columns \textit{Avg. timesteps in lost state} in Table~\ref{table:res-relocv2}. Furthermore, the proposed approach, in average creates less local maps than the \ac{DBoW2} approach as shown in columns \textit{Avg. local maps} in Table~\ref{table:res-relocv2}. This shows that a richer description for the keyframes aids the overall system, by spending less time in the lost state. If a lost state persists, it is more likely to initialize a new local map that can lead to inaccuracies in pose estimation. 
In the following lines, the performance of the proposed relocalization method is discussed independently for each of the datasets. In the \textit{MH} dataset, the proposed relocalization method shows good performance, spending only three timesteps in lost state and creating one local map. This leads to no recovery failures. In contrast the \ac{DBoW2} relocalization method spent four times as many time steps in the lost state and generated more local maps. The \ac{ATE} for the proposed approach is $0.337$ m, about $39\%$ less in comparison to the \ac{DBoW2} approach. This indicates higher positional accuracy. 
In the \textit{LiM} dataset, a similar performance is observed where the proposed relocalization method on average, spent eight timesteps in lost state and created three Local maps. In comparison, \ac{DBoW2} relocalization method created five local maps, requiring $13$ time steps to recover global pose. 

However, in sequences from \textit{VicR} and \textit{FR2PS} datasets, the performance between the two relocalization methods is similar. 
These datasets feature sequences of a drone and a ground robot surveying two indoor rooms filled with obstacles.
For the \textit{VicR} dataset, the average number of timesteps in the lost state is seven, the number of new local maps generated is three, and the \ac{ATE} score is $0.154$~m when the proposed approach is used. Comparatively, for the DBoW2 approach, $10$ timesteps are spent in the lost state, three local maps are created, and the \ac{ATE} score is $0.173$~m. 
For the \textit{FR2PS}, the numbers were $14$ timesteps, five Local maps, $0.269$ m \ac{ATE} score for the proposed. For \ac{DBoW2} approach they were $15$ timesteps, five Local maps and $0.327$ m \ac{ATE} score respectively. 
In the sequence \textit{FR2PS3}, a prolonged mapping failure towards the end of the sequence prevents recovery of global position by any method. Consequently, ORB-SLAM3 initiates a new \textit{Atlas}, leading to deletion of the entire accumulated map.

This shows that the proposed relocalization method had inherited the limitation of \ac{BA} based pose-graph optimization method that is known to fail ill-conditioned scenarios. 
Such cases are seen in the \textit{VicR} and \textit{FR2PS} datasets that show highly dynamic environments, and sometimes uncontrolled movements of the camera. However, even though both methods generated the same average number of local maps, the proposed method had a lower trajectory error in comparison to \ac{DBoW2} method.

\begin{figure*}
    \centering
    \begin{subfigure}[b]{0.32\textwidth}
    \includegraphics[width=\columnwidth]{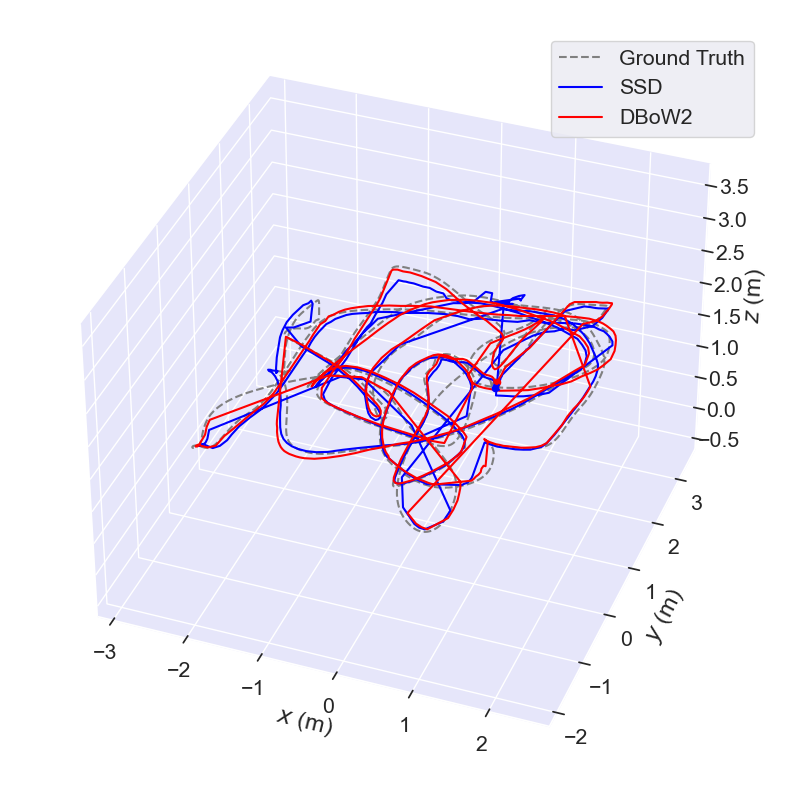}
        \caption{\textit{VicR} sequence}
        \label{fig:vicon_traj}
    \end{subfigure}
    \begin{subfigure}[b]{0.32\textwidth}
        \includegraphics[width=\textwidth]{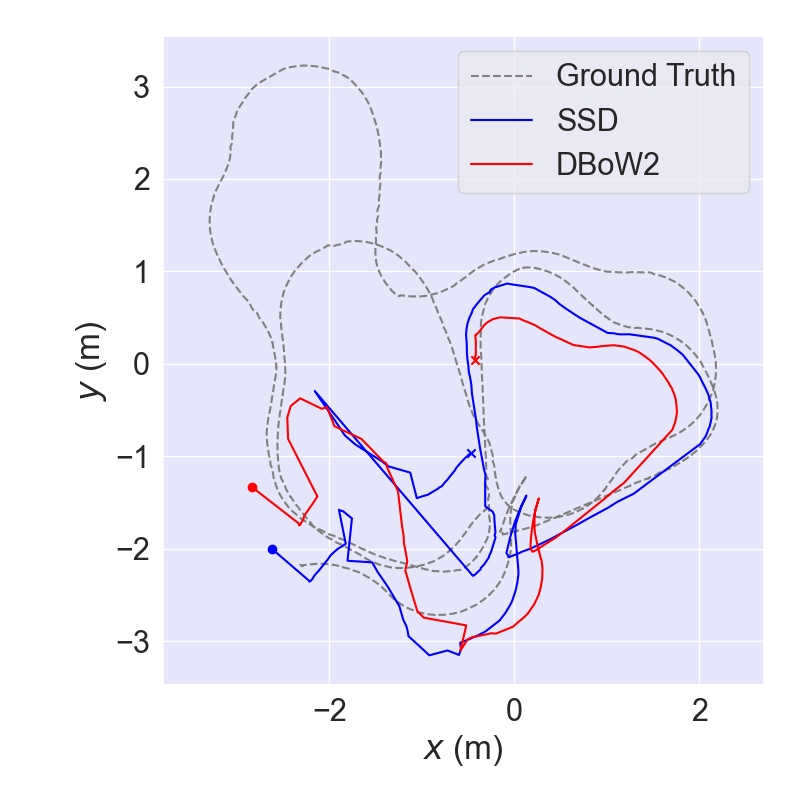}
        \caption{\textit{FR2PS} sequence}
        \label{fig:tum_traj}
    \end{subfigure}
    \caption{\small{Trajectory estimates with ORB-SLAM3 containing the modified relocalization method (blue) and \ac{DBoW2} version (red). Where available, ground-truth are shown in grey.}}\label{fig:traj_estimates}
    \label{fig:three_traj}
\end{figure*}

\figurename~\ref{fig:vicon_traj} shows the performance of the system on the \textit{V102} image sequence from \textit{VicR} dataset. In this case the robotic system visited the same area several times. The pose estimation generated by both the proposed and the \ac{DBoW2} approaches show good agreement with the ground-truth.  
%
%
\figurename~\ref{fig:tum_traj} shows the pose estimation for the \textit{FR2PS1} image sequence from the \textit{FR2PS} dataset, where both relocalization methods struggled to recover global poses. 
This can be seen from the discrepancies between the ground-truth trajectory and the estimated ones. This is due to the sudden motion of the agent carrying the camera, an ill-conditioned scenario in which \ac{BA} has limitations in recovering global pose~\cite{younes2017}.

The limitation of \ac{BA} inherited by the proposed relocalization method may be solved using techniques such as 
Object SLAM \cite{yang2019cubeslam, zhou2022point-plane-object-slam}. 
These methods are free from the 
requirement of the \ac{BA} method that requires sufficient overlap in keypoints to map points observations between the query and candidate keyframes to optimize the camera's pose. Testing these methods are beyond the scope of this paper. 

%% file: sections/Conclusions.tex
\section{Conclusion}
\label{sec:conclusions}

This paper proposes a new multimodal keyframe descriptor 
based on the semantic information of objects detected in the scene and camera poses calculated based on monocular images.  
A novel \ac{KPR} method is developed to select appropriate keyframe candidates from the pose graph using a three stage filtering algorithm. 
These two components, coupled with a \ac{PnP} solver, create a new short-term relocalization pipeline within the framework of the ORB-SLAM3. 
Across $18$ GPS-denied image sequences collected with drones and ground robots, 
the proposed approach has exceeded the performance of the bag-of-words method; this being validated through a quantitative analysis. Future work will look at the integration of 3D object pose and the information theory to expand the proposed method into a multi-agent framework.